\documentclass[10pt,twocolumn,letterpaper]{article}

\usepackage{iccv}
\usepackage{times}
\usepackage{epsfig}
\usepackage{graphicx}
\usepackage{amsmath}
\usepackage{amssymb}
\usepackage{microtype}
\usepackage{subfigure}
\usepackage{booktabs}
\usepackage{multirow}
\usepackage[table,xcdraw]{xcolor}
\usepackage{makecell}
\usepackage[accsupp]{axessibility}  

\usepackage[breaklinks=true,bookmarks=false]{hyperref}

\iccvfinalcopy 

\def\iccvPaperID{6310} 

\ificcvfinal\fi

\begin{document}

\title{RLSAC: Reinforcement Learning enhanced Sample Consensus\\ for End-to-End Robust Estimation}


\author{Chang~Nie\textsuperscript{\rm 1}, Guangming~Wang\textsuperscript{\rm 1}, Zhe~Liu\textsuperscript{\rm 2}$^{*}$, Luca~Cavalli\textsuperscript{\rm 3}, Marc~Pollefeys\textsuperscript{\rm 3,4}, Hesheng~Wang\textsuperscript{\rm 1}\thanks{ Corresponding Authors. 
		\indent~ The first two authors contributed
		equally.
	}\\
	{\textsuperscript{\rm 1}Department of Automation, Key Laboratory of System Control}\\
{and Information Processing of Ministry of Education, Shanghai Jiao Tong University}\\
    {\textsuperscript{\rm 2} MoE Key Lab of Artificial Intelligence, AI Institute, Shanghai Jiao Tong University}\\
    {\textsuperscript{\rm 3}
		Department of Computer Science, ETH Zürich}
     {\textsuperscript{\rm 4}
		Microsoft Mixed Reality and AI Zürich Lab}  ~~
 \\
	\small{\texttt{\{changnie,wangguangming,liuzhesjtu,wanghesheng\}@sjtu.edu.cn}} \\
	\small{\texttt{lcavalli@ethz.ch}}\qquad \small{\texttt{marc.pollefeys@inf.ethz.ch}}
}

\maketitle
\ificcvfinal\thispagestyle{empty}\fi

\begin{abstract}
Robust estimation is a crucial and still challenging task, which involves estimating model parameters in noisy environments. Although conventional sampling consensus-based algorithms sample several times to achieve robustness, these algorithms cannot use data features and historical information effectively. In this paper, we propose RLSAC, a novel Reinforcement Learning enhanced SAmple Consensus framework for end-to-end robust estimation. RLSAC employs a graph neural network to utilize both data and memory features to guide exploring directions for sampling the next minimum set. The feedback of downstream tasks serves as the reward for unsupervised training. Therefore, RLSAC can avoid differentiating to learn the features and the feedback of downstream tasks for end-to-end robust estimation. In addition, RLSAC integrates a state transition module that encodes both data and memory features. Our experimental results demonstrate that RLSAC can learn from features to gradually explore a better hypothesis. Through analysis, it is apparent that RLSAC can be easily transferred to other sampling consensus-based robust estimation tasks. To the best of our knowledge, RLSAC is also the first method that uses reinforcement learning to sample consensus for end-to-end robust estimation. We
release our codes at \url{https://github.com/IRMVLab/RLSAC}.
\end{abstract}

\section{Introduction}
\begin{figure}[t]
  \centering
   \includegraphics[width=1.0\linewidth]{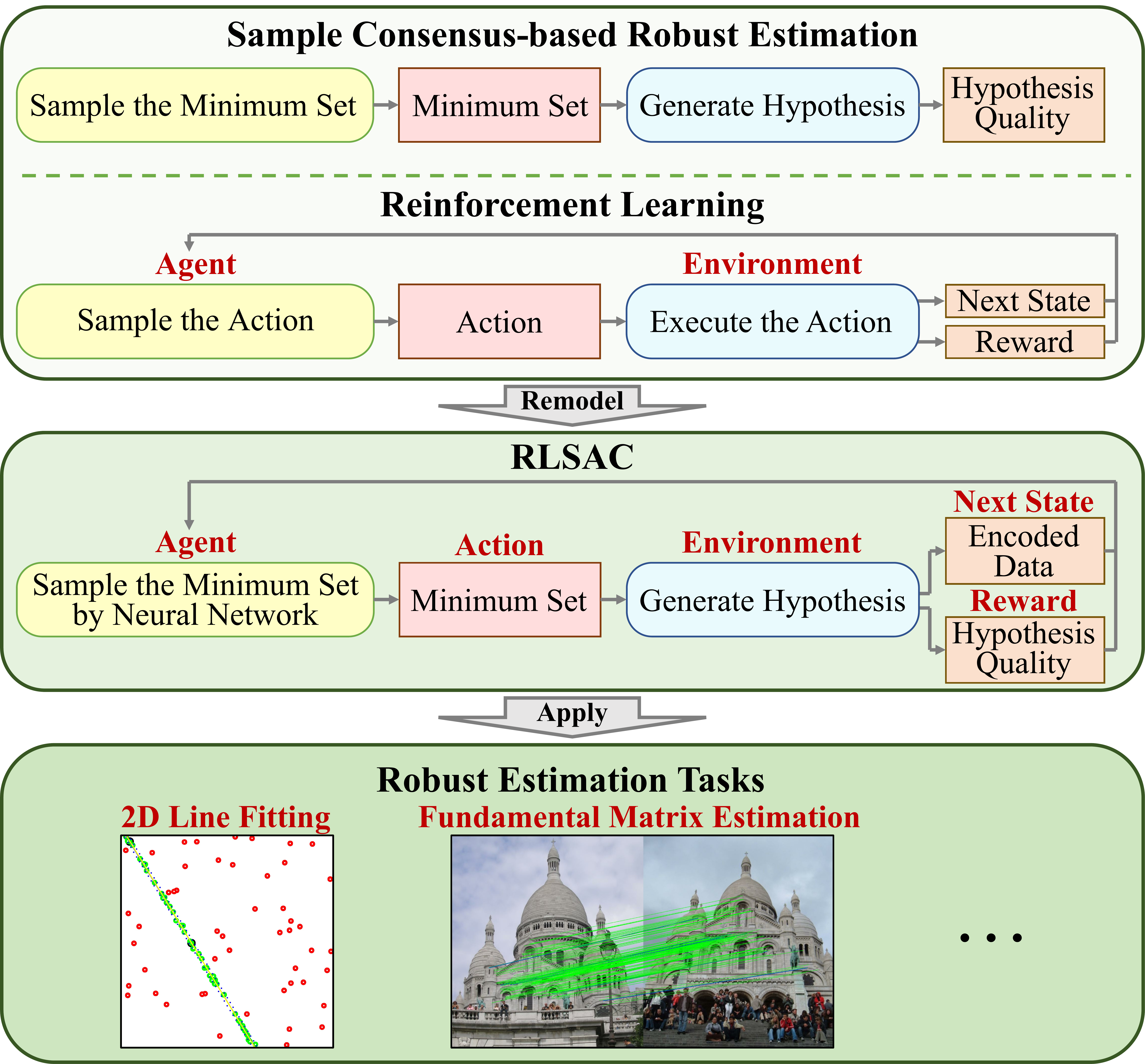}

   \caption{\textbf{RLSAC remodel the sampling consensus process.} By modeling the sampling consensus as a reinforcement learning process, RLSAC can achieve end-to-end learning on various robust estimation tasks.}
   \label{fig:overall}
\end{figure}
As a fundamental module in computer vision, robust estimation is crucial for many tasks, such as camera pose estimation \cite{camposeco_hybrid_2018, wang20223d, wang2022efficient, wang2021pwclo}, motion segmentation \cite{torr_mlesac_2000, jiang20223d, wang2022unsupervised, wang2021unsupervised}, short and wide baseline matching \cite{mishkin_mods_2015, wang2022fusionnet}, plane fitting \cite{hofer_3d_2005}, and line fitting \cite{chin_robust_2009, li_consensus_2009}. However, it is still difficult to exclude disturbances while estimating accurate models. To address this issue, sampling consensus-based algorithms are widely used, which are represented by the RANdom SAmple Consensus (RANSAC) \cite{fischler1981random} algorithm. It first samples the minimum set required for the task, \emph{e.g.}, a minimum set size of 2 for 2D line fitting. Then, the hypothesis is solved by the minimum set. Next, all data is divided into inliers and outliers, according to their residuals to the hypothesis. Finally, the above process is repeated and the best hypothesis is selected based on the highest inlier ratio. RANSAC can provide strong robustness and generalization, but as the outlier rate increases, the probability of sampling inliers decreases. As a result, the performance of RANSAC degrades rapidly. This is because RANSAC samples each data evenly, regardless of data features that can be used for classifying inliers and outliers. Additionally, the non-differentiable deterministic random sampling of RANSAC also limits it to be integrated into the learning-based pipeline.


Since sampling the minimum set from the data is quite similar to the process of sampling an action from the action space in reinforcement learning \cite{haarnoja_soft_2018, christodoulou_soft_2019}, the sampling consensus can be integrated into the reinforcement learning framework. Sampling in reinforcement learning can be achieved through a neural network, which extracts the data features. In addition, the reward from the environment can be used to train the reinforcement learning framework without differentiation. Therefore, to learn from data features and avoid differentiating, we propose the RLSAC: Reinforcement Learning enhanced SAmple Consensus for end-to-end robust estimation. As shown in Figure \ref{fig:overall}, RLSAC regards sampling consensus as the process of interaction between the agent and environment in reinforcement learning. Specifically, the agent uses a neural network to sample the minimal set from the data as an action. The environment then performs model generation and evaluation based on the action and outputs the next state, which is used in the next iteration.

However, designing appropriate reward and state are very important and challenging in reinforcement learning. To achieve reinforcement learning enhanced sampling consensus, RLSAC proposes new state transition and reward modules. Specifically, the state is encoded by augmenting the original data features with memory features, including the current action, data residuals, and historical information. When the state is input into the agent, these features can provide more information about the quality of the previous action. This allows RLSAC to gradually explore a better hypothesis by utilizing this memory information.

Additionally, the evaluation result of the generated hypothesis can be used as the reward signal to train the neural network without differentiation. The reward signal enables the learning-based sampling consensus for end-to-end robust estimation. Furthermore, the evaluation result is the feedback from the downstream task. Thus, the neural network can learn to effectively use the data features and optimize the output to meet the requirements of the downstream task.

Moreover, instead of directly predicting the final result from the data in one shot \cite{truong2022unsupervised}, RLSAC employs multiple episodes, each containing several sampling processes. In addition, RLSAC performs one random sampling at the beginning of each episode to form the initial state. This approach preserves the robustness of multiple random sampling and provides basic performance for RLSAC. Besides, RLSAC can be extended to other robust estimation tasks since it is not restricted to any specific task.

The proposed RLSAC is tested on two classic robust estimation tasks, which are the 2D line fitting task and the fundamental matrix estimation task. The experimental results show that RLSAC achieves great performance.

Our main contributions are as follows:
\vspace{-0.1cm}
\begin{itemize}
        \vspace{-0.1cm}
	\item We propose RLSAC: a novel Reinforcement Learning enhanced SAmple Consensus framework for end-to-end robust estimation. It learns data features to sample the minimum set. RLSAC retains the robustness of the multiple sampling process, while the initial random sampling can provide the basic performance.

        \vspace{-0.2cm}
	\item RLSAC proposes an approach for state encoding, which includes both current and historical information. This enables the agent to assess the quality of the previous actions and gradually explore better hypotheses. RLSAC is trained unsupervised using the reward function, which avoids differentiating the sampling process and achieves end-to-end robust estimation.

         \vspace{-0.2cm}
	\item RLSAC is evaluated on two robust estimation tasks. The 2D line fitting task demonstrates its robustness to disturbances and effective progressive exploration capability. In the fundamental matrix estimation task, RLSAC achieves state-of-the-art performance. Furthermore, RLSAC can be easily applied to other sampling consensus-based robust estimation tasks.


\end{itemize}

\section{Related Work}

\begin{figure*}[t]
  \centering
   \includegraphics[width=0.95\linewidth]{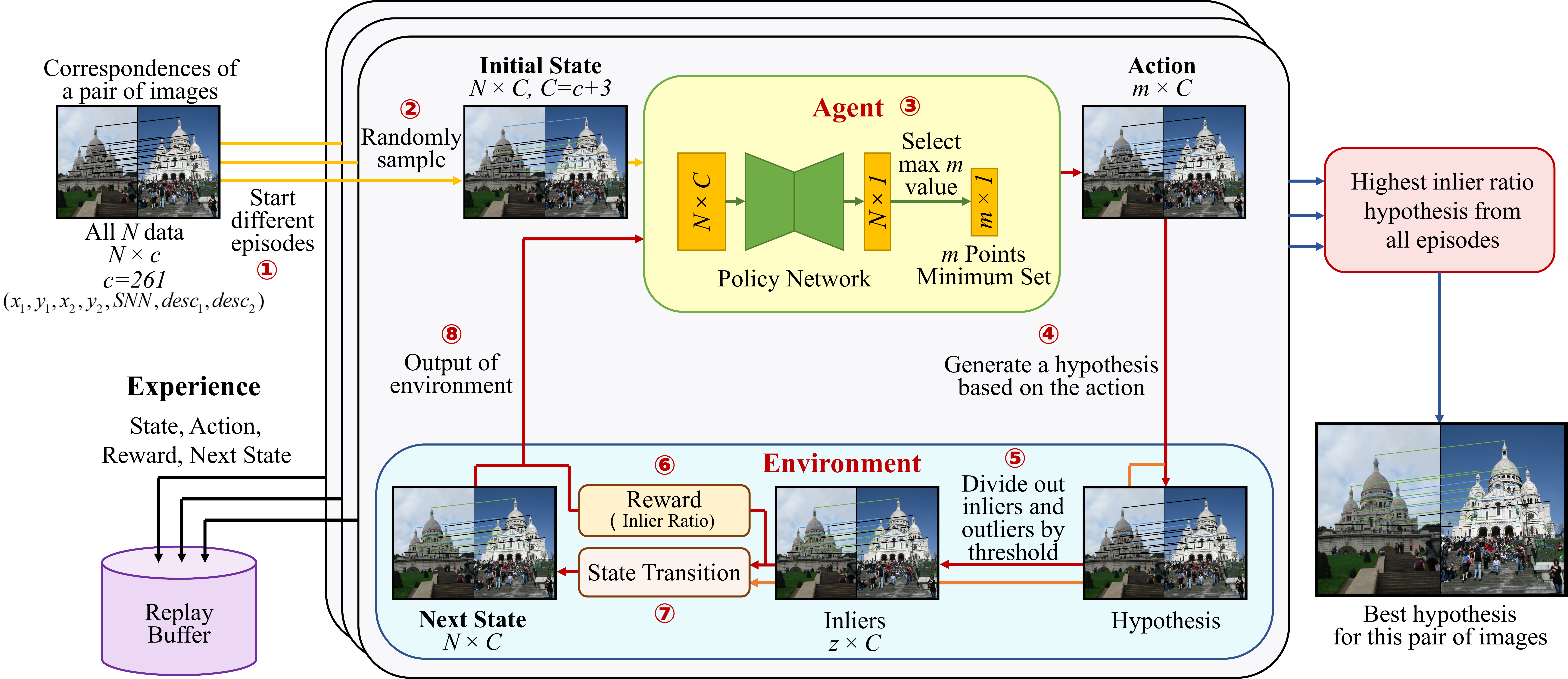}

   \caption{\textbf{The pipeline of the proposed RLSAC.} The fundamental matrix estimation problem is used as an example. The black, green, and blue lines represent outliers, inliers, and the minimum set, respectively. The yellow arrows are only used once during initialization. The red and orange arrows indicate the loop in an episode. The initial states are randomly sampled. The best hypothesis for this scene is selected by scoring all hypotheses. The collected experience is recorded in the replay buffer for training.}
   \label{fig:pipeline}
\end{figure*}

Robust estimation is a basic module for many tasks.
Although the simple repetitive random sampling strategy of RANSAC \cite{fischler1981random} is robust and generalized, it has some limitations, such as no further optimization, inefficient sampling, and non-differentiability. 


There are several methods observing that local features help to optimize the sampling result. LO-RANSAC \cite{chum2003locally} continues to sample in the inliers with a smaller threshold after sampling the best model so far, aiming to find a better hypothesis near the current one. NAPSAC \cite{torr2002napsac} uses a fixed hypersphere to acquire local data and samples from it, but this approach loses global features and may get stuck in local data. To address this, progressive NAPSAC \cite{barath_progressive_2019} gradually expands the hypersphere to extend the sampling from local to global. GC-RANSAC \cite{barath2018graph} considers spatial continuity between inliers and surrounding points, modeling the connections of the data through graph-cut to divide inliers and outliers. MAGSAC++ \cite{barath_marginalizing_2022} assesses the model quality by weighting various thresholds to reduce the sensitivity to the choice of a specific noise scale.

To improve sampling efficiency, some methods use guided sampling instead of random sampling. PROSAC \cite{chum2005matching} sorts the data based on a quality function and then samples the sorted data sequentially to improve efficiency. USAC \cite{raguram_usac_2013} integrates the advantages of various methods to achieve robustness and efficiency. To learn from the data, Yi et al. \cite{Yi_2018_CVPR}  are the first to use a neural network based on PointNet to directly classify inliers and outliers. Then, Zhang et al. \cite{zhang2019learning} improve this work by proposing pooling and unpooling blocks to learn the local context of correspondences. NG-RANSAC \cite{brachmann2019neural} uses a neural network to calculate the sample probability for each point, achieving probability-based sampling. The neural network is trained using reinforcement learning, but it did not use reinforcement learning to achieve sampling consensus and progressive exploration. Barath et al. \cite{barath_learning_2022} propose the MQ-Net to learn from residual histograms and evaluate the quality of the model. Additionally, the authors design the MF-Net to learn to reject bad minimum sets early, further improving efficiency. NeFSAC \cite{cavalli_nefsac_2022} uses a neural network to reject the motion-inconsistent and poorly-conditioned minimal samples. 

The non-differentiability of RANSAC makes it challenging to integrate into an end-to-end learning pipeline. Some methods propose the differentiable RANSAC. For example, DSAC \cite{brachmann_dsac_2017} replaces the deterministic selection process with probabilistic selection, allowing for differentiation with respect to the data. Wei et al. \cite{wei_fully_2022} achieve gradient propagation by predicting the probabilities of being inliers to guide sampling. To avoid differentiation, some methods use the loss as reward signals through reinforcement learning to train the network. Bhowmik et al. \cite{bhowmik2020reinforced} use reinforcement learning to train the feature point detection network but not for the sampling consensus process. Truong et al. \cite{truong2022unsupervised} iteratively delete points with reinforcement learning, seeking the maximum consensus model that meets the threshold. However, the authors encode the attributes of the points into the state, but does not include long-range historical information or the position of the hypothesis in the state space. In addition, this method does not include the sampling consensus process and may overlook better solutions.

\section{Method}
\subsection{Problem Formulation}
Robust estimation can be considered as generating a good hypothesis $h$ given a set of data $\chi  = \left\{ {{{\rm{x}}_i}} \right\}_{i = 1}^N$, which may be disturbed by noise. For instance, the hypothesis $h$ could represent the parameters of a 2D line, and $\chi$ could be all the points in the 2D plane. Or $h$ could be the fundamental matrix that represents the epipolar geometry of a pair of images, and $\chi$ could be all the correspondences. Similarly, the data contained in ${{\rm{x}}_i}$ varies with the task. As a commonly used robust estimation method, the sampling consensus method samples $n$ minimum sets $\mathcal{M}$ from $\chi$. The size $m$ of a minimum set varies depending on the task. For example, when fitting a 2D line, the size $m=2$.
Then, these minimum sets $\mathcal{M}$ are solved by the minimum solver $S$ to output hypotheses:
\vspace{-0.2cm}
\begin{equation}
H = \left\{ {S\left( {{\mathcal{M}_j}} \right)\left| {{{\mathcal{M}_j}} \in M,j = 1,2,...,n} \right.} \right\}.
\end{equation}

\vspace{-0.2cm}
Next, the hypotheses $H$ are evaluated using a scoring function $f$. The residuals of all data points $\chi$ can be computed and used to calculate the inlier ratio, which is commonly used as a metric for the hypothesis quality \cite{fischler1981random}. Finally, the hypothesis with the highest score is chosen as the best hypothesis $h_{Best}$:
\vspace{-0.2cm}
\begin{equation}
{h_{Best}} = \mathop {\arg \max }\limits_{h \in H} f\left( {h,\chi } \right).
\label{eq:hbest}
\end{equation}
By solving the problem using several minimum sets $\mathcal{M}$, the model can achieve robust estimation, which is less sensitive to outliers and can lead to more accurate results.

The RANSAC \cite{fischler1981random} algorithm performs random sampling to sample the minimum sets $\mathcal{M}$, which cannot fully exploit the data features. As shown in Figure \ref{fig:overall}, One alternative is to consider the sampling of a minimum set as an action taken by an agent based on the current state, within the reinforcement learning framework:
\vspace{-0.2cm}
\begin{equation}
a_{t+1} \sim \pi_\phi\left(a_t \mid s_t\right).
\end{equation}
Here, the policy network $\pi_\phi$ depends on the weights $\phi$. The action $a_{t+1}$ is sampled by the policy $\pi_\phi$ from the state $s_t$ and $a_{t}$ at time $t+1$, which can also be viewed as the process of sampling a minimum set $\mathcal{M}_j$ from all data $\chi$. So the Eq. \ref{eq:hbest} can be rewritten that the best hypothesis is selected by a reinforcement learning enhanced sample consensus method:
\vspace{-0.2cm}
\begin{equation}
h_{Best} = \mathop {\arg \max }\limits_{t = 1,2,...,j} f\left( { S\left( {{a_t}} \right) ,\chi } \right).
\end{equation}

\vspace{-0.2cm}
In addition, the policy $\pi_\phi$ is a trainable neural network, and the state $s_t$ contains the features of the data $\chi$. Therefore, the minimum set can be selected based on sufficient learned knowledge of the data features, rather than being chosen randomly. Furthermore, the evaluation result $f\left( { S\left( {{a_t}} \right) ,\chi } \right)$ of the minimum solver $S$ can serve as the reward in reinforcement learning to achieve unsupervised learning.

\subsection{System Framework}
With the problem formulation of modeling the process of sampling consensus as a reinforcement learning problem, we introduce RLSAC, which is illustrated in Figure \ref{fig:pipeline}. Although the fundamental matrix estimation task is used as an example, the pipeline is also applicable to other robust estimation tasks. Firstly, RLSAC initiates multiple episodes for the correspondences of a pair of images $\chi$ at \textcircled{\raisebox{-0.7pt}{1}}. An episode contains many steps. At the start of each episode, a random sampling of the minimum set $\mathcal{M}_0$ is performed to generate the initial state ${s_0}$ through state transition at \textcircled{\raisebox{-0.7pt}{2}}. Specifically, the initial state ${s_0}$ is obtained by concatenating each data point with the memory features, which are consisting of action, residual, and historical features (see Section \ref{transition}).

Next, the initial state ${s_0}$ is input into the sampling consensus loop. The agent receives the current state ${s_t}$ and feeds it into the policy network $\pi_\phi$ at \textcircled{\raisebox{-0.7pt}{3}}, as shown in Section \ref{agent}. The network generates a probability for each data point. The $m$ points with the highest probabilities are selected as the minimum set $\mathcal{M}_t$, instead of random sampling.

Then, the minimum set $\mathcal{M}_t$ serves as the action $a_t$ for the agent, which generates a hypothesis ${h_t}$ in the environment at \textcircled{\raisebox{-0.7pt}{4}}, as shown in Section \ref{environment}. The residuals of points to the hypothesis are compared with a threshold to classify points into inliers and outliers at \textcircled{\raisebox{-0.7pt}{5}}. The ratio of inliers is utilized as the reward $r_t$ for the current action $a_t$ to train the policy network $\pi_\phi$ at \textcircled{\raisebox{-0.7pt}{6}}. Additionally, the action, inliers, and residuals are used for the state transition to generate the next state ${s_{t+1}}$ at \textcircled{\raisebox{-0.7pt}{7}}, as shown in Section \ref{transition}. Finally, the environment outputs the next state ${s_{t+1}}$ and the reward $r_t$, which are used by the agent to start the next step at \textcircled{\raisebox{-0.7pt}{8}}.

Furthermore, the state ${s_t}$, action $a_t$, reward $r_t$, and the next state ${s_{t+1}}$ in every step are collected as experiences in the replay buffer \cite{christodoulou_soft_2019} for training. After all episodes are complete, the hypothesis with the highest inlier ratio across all episodes is output as the best hypothesis ${h_{Best}}$ for this pair of images.

\subsection{Sampling Minimum Set by Agent}
\label{agent}
The agent receives the state  ${s_t} \in {\mathbb{R}^{N \times C}}$, where $N$ is the number of the correspondences of a pair of images. The channel $C=c+3$, with $c$ denoting the dimension of data features and $3$ representing the dimension of memory features (see Section \ref{transition}). In addition, the first hypothesis $h_0$ of each episode is generated using a randomly sampled minimum set, which provides a basic performance for RLSAC.

Then the state ${s_t}$ is fed into the policy network $\pi_\phi$. To achieve permutation invariance for the input data, RLSAC uses edge convolution (EdgeConv) from the DGCNN \cite{wang_dynamic_2019} as the basic module to establish the policy network $\pi_\phi$. The EdgeConv can model the interrelationship of correspondences by graph neural networks. It can extract features from neighboring nodes in a graph and aggregate them into a central node. The policy network $\pi_\phi$ extracts data features and calculates the probabilities by softmax for each point of being in the minimum set. The probability can be interpreted as the probability of obtaining a greater return on the action for the long term, rather than the probability of achieving the best result in the current state. Consequently, the $m$ points with the highest probability are selected as the minimum set $\mathcal{M}_t$ in the current state ${s_t}$. Importantly, all previously used minimum sets are recorded to avoid selecting a duplicate minimum set. Thus, when a new minimum set is selected, it is checked whether it has been used. If the new minimum set has already been used, then another $m$ data set is selected following the probability, which will be checked as well. Until the unused data set is found, it is output as the minimum set. Ultimately, the minimum set is as the action $a_t \in {\mathbb{R}^{m \times C}}$ output by the agent.

\subsection{Evaluating Hypothesis in Environment}
\label{environment}
The environment generates a hypothesis $h_t$ based on the received action $a_t$ from the agent. Next, the residuals $R$ are calculated for all data $\chi$ with respect to the hypothesis:
\begin{equation}
R_t = \left\{ {D\left( {{x_i},{h_t}} \right)\left| {{x_i} \in \chi ,i = 1,2,...,N} \right.} \right\},
\label{eq:rt}
\end{equation}
where the function $D$ is for residual calculation. The scalar valued residuals $R$ are calculated at all data $\chi$.
With the pre-defined threshold, the data $\chi$ is divided into inliers $I_t \in {\mathbb{R}^{z \times C}}$ and outliers $O_t \in {\mathbb{R}^{\left( {N - z} \right) \times C}}$ based to the residuals.

In the sampling consensus methods, the inlier ratio commonly represents the quality of the hypothesis. Since the hypothesis $h_t$ is generated by the action $a_t$ in RLSAC, the inlier ratio can be considered as the reward $r_t$ of the current action $a_t$ to train the policy network $\pi_\phi$ without ground truth.

With the hypotheses related-rewards, RLSAC can update weights $\phi$ of the policy network $\pi$ without differentiating the sampling process. By modeling sampling consensus as an interactive process in reinforcement learning, RLSAC achieves end-to-end robust estimation while avoiding differentiating.

\begin{figure}[t]
  \centering
   \includegraphics[width=0.95\linewidth]{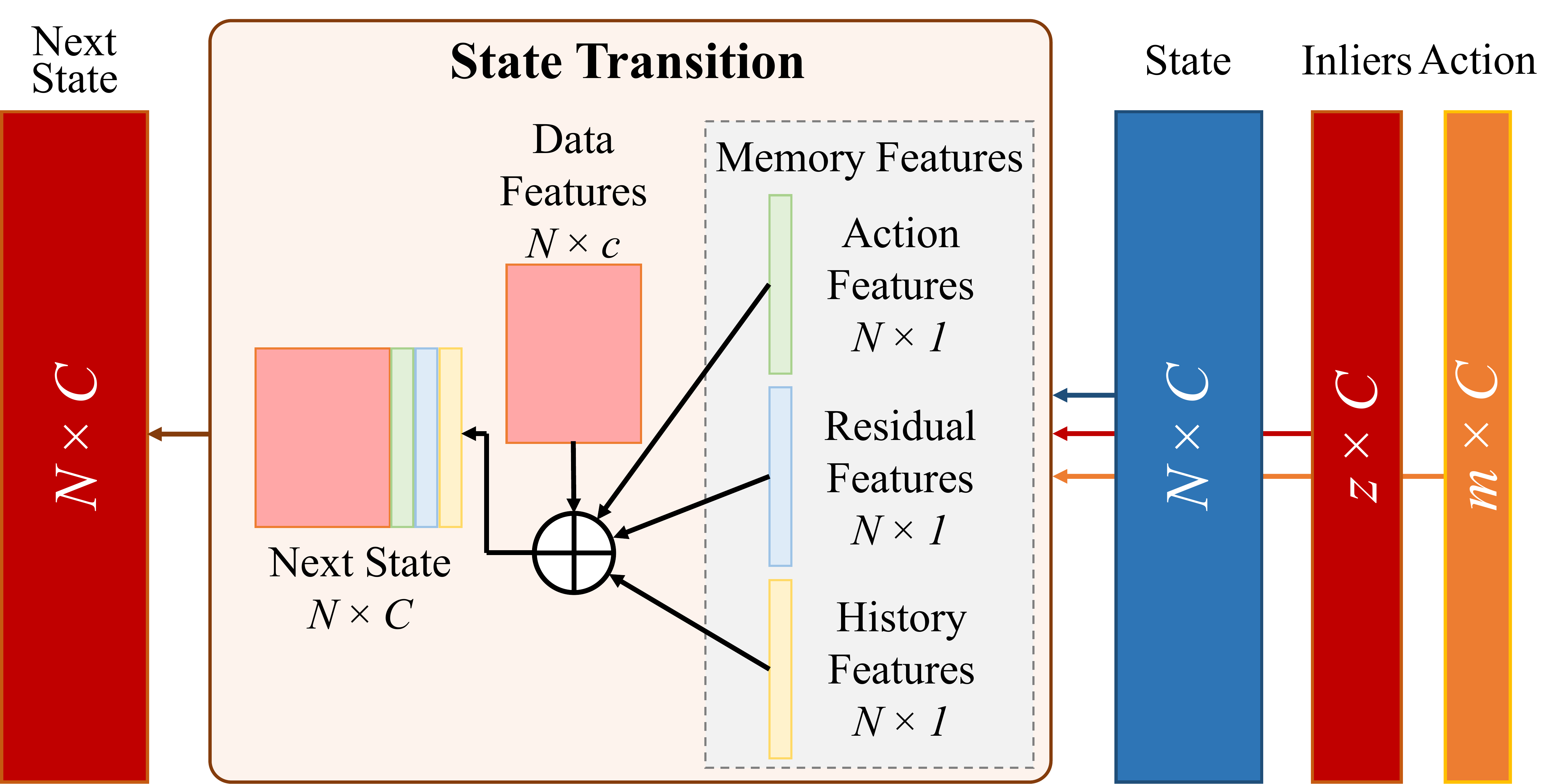}

   \caption{\textbf{The state transition in RLSAC.} To form the next state, the data features are added with the memory features, which contain action features, residual features, and historical features.}
   \label{fig:transition}
\end{figure}

\subsection{State Transition}
\label{transition}
The state transition module allows RLSAC to encode the previous state and action, enabling RLSAC to explore the state space effectively. As illustrated in Figure \ref{fig:transition}, the next state $s_{t+1}$ is obtained by concatenating the original data features $N \times c$ with memory features $\left\{ {A,R,\mathcal{H}} \right\}$, which comprise action features $N \times 1$, residual features $N \times 1$ and historical features $N \times 1$. 

The action features are derived from the current action $a_t$. If a data point $x_i$ is used in the action $a_t$, the corresponding entry in the action features ${\alpha _i}$ is set to $1$, otherwise it is $-1$:
\begin{equation}
A_t = \left\{ {{\alpha _i}} \right\},{\alpha _i} = \left\{ {\begin{array}{*{20}{l}}
1&{{\rm{if\ }}{x_i} \in {a_t}}\\
{ - 1}&{{\rm{otherwise}}}
\end{array}} \right.,i = 1,2,...,N.
\end{equation}
The action features could provide the agent with information on the current action, like the current location in the data for the next state.

The residual features are obtained from the residuals $R_t$ computed in Eq. \ref{eq:rt}. They encode the relative relationship between the hypothesis and the data, which can be thought of as the direction for the agent.

The historical features collect how often is a data point used up to the current step. They are initialized to $0$ and updated as follow: for each data point $x_i$ used in the current action $a_t$, the corresponding entry in the historical features ${\tau _i}$ is incremented by $1$:
\begin{equation}
{\mathcal{H}_t} = \left\{ {{\tau _i}} \right\},{\tau _i} = {\tau _i} + 1{\rm{\ if\ }}{x_i} \in {a_{\rm{t}}},i = 1,2,...,N.
\end{equation}
The historical features can provide the agent with information on the actions taken so far, enabling it to keep track of its path through the data.

By encoding the state in this way, the agent has access to both current and historical information, allowing it to reach the goal state. Furthermore, the memory features serve as a director of local navigation, guiding the agent to explore the data features and reach its destination gradually.

\section{Experiments}
\label{Experiments}
We evaluate the performance of RLSAC on classic tasks, including 2D line fitting and fundamental matrix estimation. The 2D line fitting task serves as a basic benchmark to assess the performance of the robust estimation algorithm and provides visualization of the sampling process. The fundamental matrix estimation task can show the performance of algorithms on complex camera pose estimation task in the real world, which is a critical task in computer vision.

\begin{table*}[t]\small 
\centering
\caption{\textbf{2D line fitting.} The mAA@$0.5^{\circ}$ and median error($^{\circ}$) on various outlier rates are reported.}
\setlength{\tabcolsep}{0.9mm}
\renewcommand\arraystretch{0.8}
\begin{tabular}{c||cc|cc|cc|cc|cc|cc|cc}
\toprule
\multirow{2}{*}{Method} & \multicolumn{2}{c|}{0.1} & \multicolumn{2}{c|}{0.2} & \multicolumn{2}{c|}{0.3} & \multicolumn{2}{c|}{0.4} & \multicolumn{2}{c|}{0.5} & \multicolumn{2}{c|}{0.6}  & \multicolumn{2}{c}{0.7} \\ \cline{2-15} 
                        & mAA ↑    & Mid. ↓       & mAA ↑       & Mid. ↓    & mAA ↑       & Mid. ↓       & mAA ↑        & Mid. ↓      & mAA ↑       & Mid. ↓       & mAA ↑       & Mid. ↓ & mAA ↑       & Mid. ↓    \\ \hline \hline \noalign{\smallskip}
RANSAC \cite{fischler1981random}    & 0.870 & 0.049 & 0.863 & 0.052 & 0.850 & 0.056 & 0.829 & 0.061 & 0.796 & 0.071 & 0.746 & 0.087 & 0.608 & 0.135 \\
Ours     & \textbf{0.875} & \textbf{0.047} & \textbf{0.874} & \textbf{0.049} & \textbf{0.872} & \textbf{0.048} & \textbf{0.865} & \textbf{0.050} & \textbf{0.858} & \textbf{0.052} & \textbf{0.845} & \textbf{0.056} & \textbf{0.824} & \textbf{0.062} \\
\hline \hline
Ours-0.5     & 0.849 & 0.053 & 0.850 & 0.055 & 0.854 & 0.053 & 0.858 & 0.052 & \textbf{0.858} & \textbf{0.052} & \textbf{0.864} & \textbf{0.050} & \textbf{0.849} & \textbf{0.054} \\\bottomrule
\end{tabular}
\label{tab:line_result}
\vspace{-0.3cm}
\end{table*}

\subsection{Settings}
RLSAC is built on the widely used reinforcement learning framework SAC-Discrete \cite{christodoulou_soft_2019}. As shown in Figure \ref{fig:pipeline}, the collected replay buffer is retrieved for off-policy to update the actor and critic network in reinforcement learning.

Since RLSAC uses the inlier ratio as the reward for training, it is an unsupervised method. To explore more images, only one episode with multiple steps is collected per image pair during training. The framework is trained for 100 epochs. During training, the episode termination conditions are set as follows: (i) if the number of inliers is unchanged for $\kappa=2$ steps; (ii) if the inlier ratio does not exceed the maximum inlier ratio in the episode for $\varsigma=3$ steps; (iii) if the maximum number of steps $\psi=15$ is reached. During testing, only the third condition is valid, and each pair of images can be tested in $\nu$ episodes.

As with the standard reinforcement learning \cite{christodoulou_soft_2019}, RLSAC uses probabilistic sampling during training and max sampling during testing. For the EdgeConv module in the policy network, the value of k-nearest neighbors is set to $k=15$. The network architecture and additional information can be found in the supplementary material. Moreover, we have included ablation experiments to analyze the impact of specific network details, also provided in the supplementary material. Similar to the RANSAC implementation, RLSAC also employs refinement using the inliers of the final best hypothesis as the final model polishing.

The experiments are conducted on a Linux computer with an Intel i7 3.6GHz CPU and an NVIDIA RTX 2080Ti GPU. Our implementation is based on PyTorch.

\subsection{Case Study 1: 2D Line Fitting}
The 2D line fitting task is a basic problem in robust estimation, which can visualize the process of robust estimation and assess the robustness quantitatively to different outlier rates. In this task, each data point contains only the coordinates, thus all data points can be expressed as:
$\chi  = \left\{ {\left[ {{x_i},{y_i}} \right]\left| {\ i = 1,2,...,N} \right.} \right\}$, where $\chi$ represents the data features $N \times c$, with $c=2$.
Since two points are sufficient to determine a 2D line, the agent can generate a hypothesis by selecting two points from the set of all data points as the minimum set $N \times m$, where $m=2$.
For comparison, the RANSAC \cite{fischler1981random} algorithm is evaluated with the same task.

To evaluate the performance of RLSAC in the 2D line fitting task, we synthesize $N=100$ data points for both training and testing. Specifically, a ground truth 2D line is randomly generated in a $10 \times 10$ picture. True inliers are then randomly generated on the line based on the set outlier rate. Next, the true inliers are uniformly disturbed within a range of 0.1 around the line. Thus, the inlier threshold could be set to $\varepsilon=0.1$. Additionally, true outliers are randomly scattered throughout the picture. Moreover, it is possible for true outliers to be located within the inlier region, as is often the case in real-world scenes. 

For accurately evaluating and comparing the performance of different methods, we adopt the mean Average Accuracy (mAA) metric in \cite{barath_learning_2022}. The angular difference between the estimated line and the ground truth line is used as the error metric, which is used to calculate the mAA metric with a tolerance threshold of $0.5^{\circ}$.

The performance of RLSAC and RANSAC at different outlier rates with 150 iterations is evaluated and the results are presented in Table \ref{tab:line_result}. When the outlier rate is lower than 0.5, RANSAC can achieve a similar performance to RLSAC. This is because the repeated random selection of the minimum set can achieve good performance in simple low outlier rate scenes. However, the performance of RANSAC degrades more quickly than that of RLSAC as the outlier rate increases. This suggests that RLSAC can explore hypotheses closer to the ground truth in noisy scenes.

As the outlier rate increases beyond 0.5, RLSAC can still maintain a low error and high mAA score. This indicates that RLSAC is more robust to disturbances, providing a more stable performance compared to RANSAC.

\begin{figure}[t]
  \centering
   \includegraphics[width=0.95\linewidth]{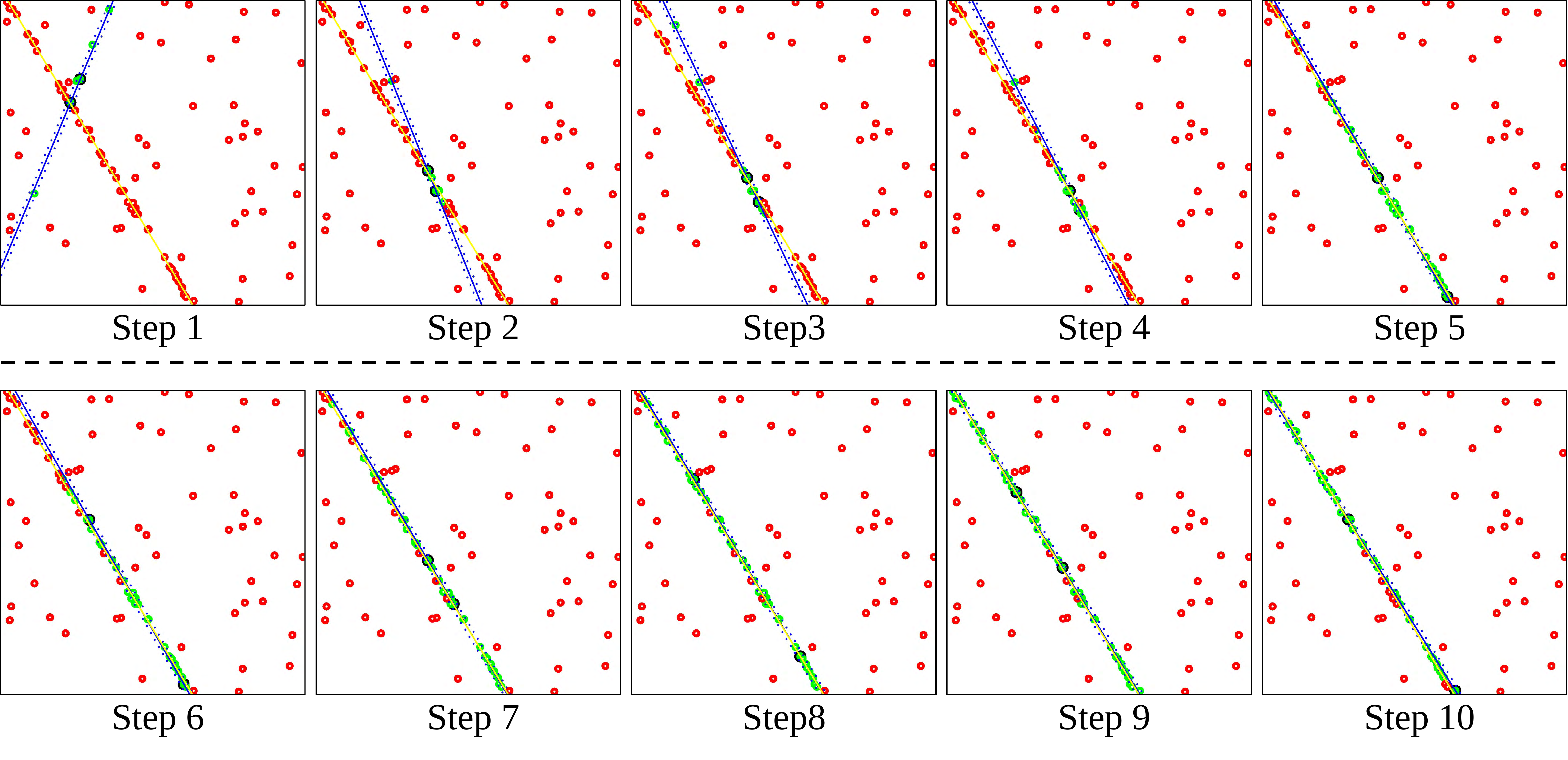}

   \caption{\textbf{The qualitative results of RLSAC on 2D line fitting.} The hypothesis converges to the ground truth as the number of steps increases. The green points represent inliers, while the red points represent outliers. The sampled minimum set points are denoted by black edges. The ground truth is represented by a yellow line, while the hypothesis and inlier threshold are represented by blue and dashed lines respectively.}
   \label{fig:line}
\end{figure}

The qualitative results of RLSAC are shown in Figure \ref{fig:line}, which demonstrates its ability to quickly find the better hypothesis even from a bad initial state. This means that the memory features can serve as a director of local navigation, allowing RLSAC to move towards a better state through outputting actions. Moreover, once RLSAC finds a high inlier ratio hypothesis that is close to the ground truth, it continues to explore the local state to further improve the result, rather than randomly transitioning to other states.

The results of the 2D line fitting task show that RLSAC can effectively utilize both data features and memory features, maintaining robust and stable performance in noisy environments. In addition, it can gradually explore a better hypothesis with the help of memory features.

\subsection{Case Study 2: Fundamental Matrix Estimation}
The fundamental matrix estimation is a crucial robust estimation task in computer vision, which solves the fundamental matrix by correspondences in a pair of images. In this study, we use the data and settings in CVPR tutorial \textit{RANSAC in 2020} \cite{barath2020ransac}, where the correspondences are detected by RootSIFT \cite{arandjelovic_three_2012} and matched by the nearest neighbor. The training data comprises 12 scenes, each with 100k image pairs, while the test data includes 2 scenes, with 4950 image pairs. The dataset uses Second Nearest Neighbor (SNN) ratio as the matching score. The correspondences are sorted in descending order by SNN. Then the top $N=150$ SNN correspondences are selected. Next, each point extracts a 128-dimensional descriptor $desc$ through the SIFT algorithm. Thus, the data points are: $\chi  = \left\{ {\left[ {x_1^i,y_1^i,x_2^i,y_2^i,SNN^i,desc_1^i,desc_2^i} \right]\left| {i = 1,2,...,N} \right.} \right\} \in {\mathbb{R}^{N \times c}}$, where $c=261$. The 8-point solver is used to solve the hypotheses \cite{longuet1981computer}. The inlier threshold of RLSAC is set to $\varepsilon=4$. The evaluation settings follow \cite{barath2020ransac}.
To compare with other methods, RANSAC \cite{fischler1981random}, USAC \cite{raguram_usac_2013}, MAGSAC++ \cite{barath_marginalizing_2022} are tested, which use the recommended settings in \cite{barath2020ransac}.




\begin{table}[t]\small 
\centering
\caption{\textbf{Fundamental matrix estimation.} The mAA@$10^{\circ}$ and median error(${\epsilon_\textbf{R}}$ and ${\epsilon_\textbf{t}}$) of rotation and the direction of translation at 1k iterations are reported in degrees.}
\setlength{\tabcolsep}{2.0mm}
\renewcommand\arraystretch{0.9}
\begin{tabular}{c||cc|cc}
\toprule
\multirow{2}{*}{Method} & \multicolumn{2}{c|}{mAA@$10^{\circ}$ ↑} & \multicolumn{2}{c}{Median ($^{\circ}$) ↓} \\ \cline{2-5} 
                        & \textbf{R}        & \textbf{t}        & ${\epsilon_\textbf{R}}$        & ${\epsilon_\textbf{t}}$   \\ \hline \hline \noalign{\smallskip}
RANSAC\cite{fischler1981random}            & 0.644            & 0.488            & 2.307        & 5.100  \\
USAC\cite{raguram_usac_2013}               & 0.741            & 0.604            & 1.036        & 2.157   \\

MAGSAC++\cite{barath_marginalizing_2022}     & 0.753    & 0.614            & \textbf{0.924}        & 1.895   \\

Ours                    & \textbf{0.760}             & \textbf{0.622}            & 0.926            & \textbf{1.751} \\ \bottomrule
\end{tabular}
\label{tab:f_result}
\vspace{-0.2cm}
\end{table}

\begin{figure}[t]
  \centering
   \includegraphics[width=0.9\linewidth]{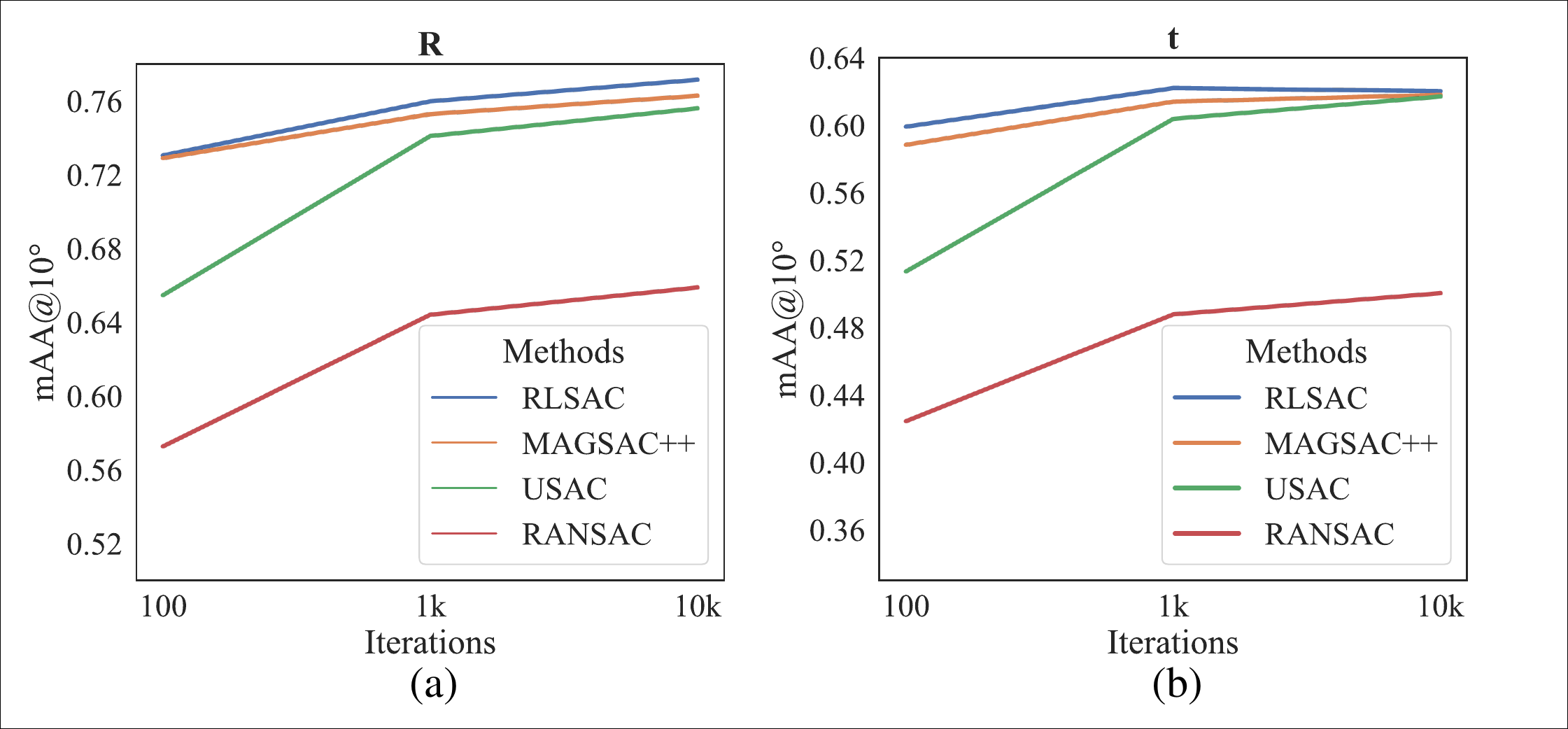}

   \caption{\textbf{The mAA@$10^{\circ}$ at different iterations.}
   The results on the fundamental matrix estimation task at different iterations.} 
   \label{fig:iterations}
\end{figure}

\begin{figure}[t]
  \centering
   \includegraphics[width=0.95\linewidth]{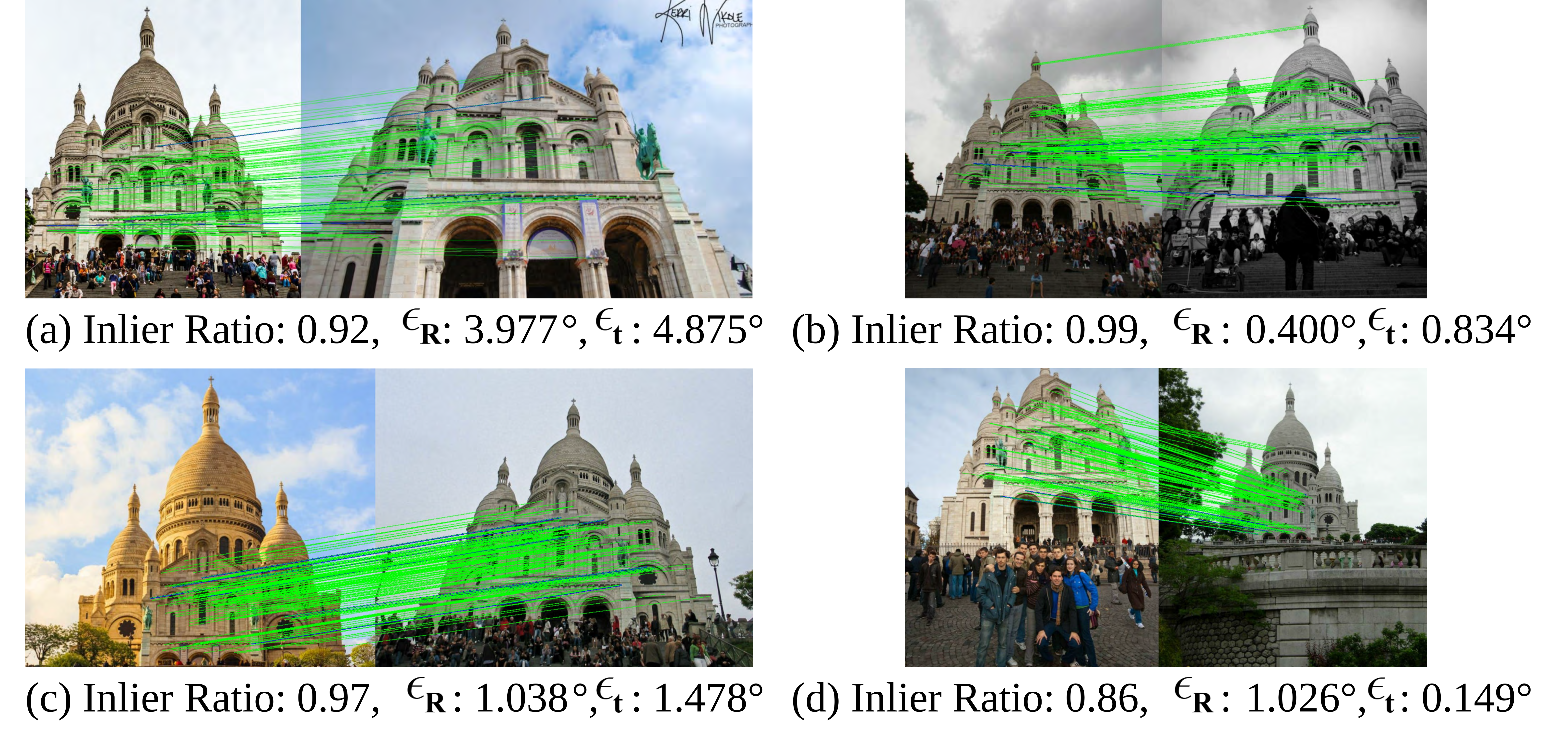}

   \caption{\textbf{The qualitative results of RLSAC on the fundamental matrix estimation task.} Inlier rate, rotation and translation errors are reported. The blue lines represent the sampled minimum set points, and the green lines represent the inliers.} 
   \label{fig:f}
\end{figure}

\begin{figure}[t]
  \centering
   \includegraphics[width=0.95\linewidth]{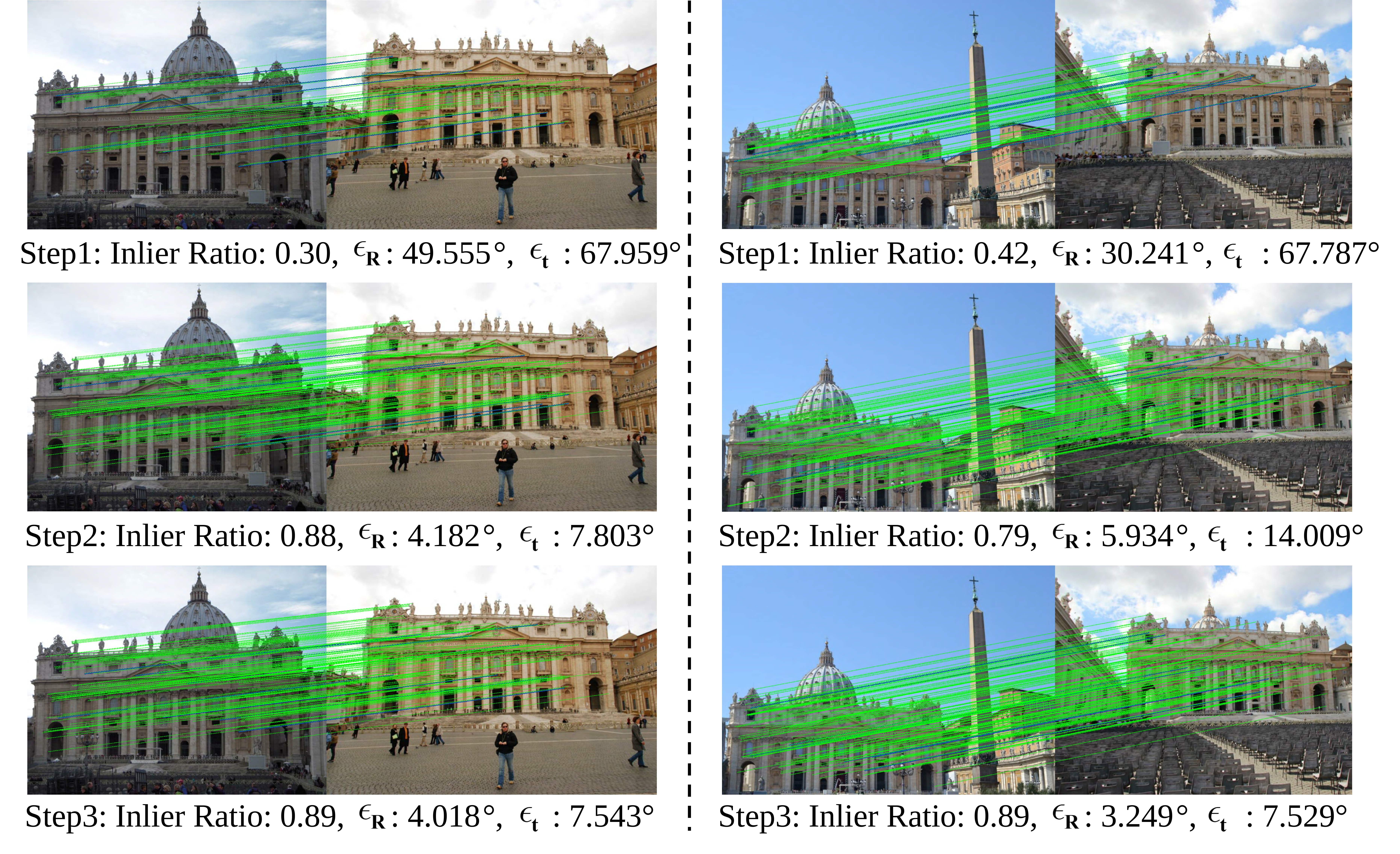}

   \caption{\textbf{The step results of RLSAC on the fundamental matrix estimation task.} The meaning of the lines and the evaluation metrics are consistent with Figure \ref{fig:f}.} 
   \label{fig:f_step}
\vspace{-0.5cm}
\end{figure}

As shown in Figure \ref{fig:iterations}, RLSAC outperforms other methods in estimating both the rotation matrix and translation vector. Remarkably, RLSAC can achieve high accuracy with only 100 iterations. Comparisons reveals that RLSAC possesses a higher upper bound in performance than MAGSAC.


The quantitative results of the methods at 1k iterations are presented in Table \ref{tab:f_result}. RLSAC achieves great performance with small errors, a little higher than MAGSAC++\cite{barath_marginalizing_2022}. Specifically, RLSAC shows high performance in estimating the direction of the translation vector compared to other methods. Since RLSAC could effectively learn features from the data to exclude noise disturbances, it has significant performance on translation vector estimation.

Figure \ref{fig:f} shows the qualitative results of the fundamental matrix estimation of RLSAC in each scene. The figure demonstrates that RLSAC selects rigid and fixed feature points on buildings as the minimum sets, which can help to find better poses. Furthermore, the step-by-step fundamental matrix estimation results of RLSAC are visualized in Figure \ref{fig:f_step}. It demonstrates a gradual increase in the inlier ratio of the hypothesis and a decrease in the rotation and translation errors. This illustrates that RLSAC can progressively explore the state space to find a better hypothesis.



\section{Ablation Study}
We perform ablation studies by modifying or removing modules to analyze their effectiveness. The experimental data and settings remain the same as in Section \ref{Experiments}.

\textbf{Robustness and Generalization of RLSAC in 2D Line Fitting:}
The \emph{Ours-0.5} in Table \ref{tab:line_result} shows the result of evaluating the robustness and generalization of RLSAC in 2D Line Fitting. We train RLSAC on data with the outlier rate of 0.5 and then test it on various outlier rates. In most cases, \emph{Ours-0.5} can outperform RANSAC.
Notably, when the outlier rate exceeds 0.5, \emph{Ours-0.5} performs even better than the model trained at that outlier rate. This is because, in scenes with high outlier rates, the point distribution is closer to a uniform distribution, which lacks the features of a dense distribution. As a result, the model $RLSAC_{High}$ trained on high outlier rates may not have learned the dense distribution in low outlier rates. This results in a mediocre performance of $RLSAC_{High}$ in low outlier rates scenes. Conversely, scenes of low outlier rates exhibit both sparse and dense point distributions in different regions, providing more diverse features for the model $RLSAC_{Low}$ to learn. Consequently, the $RLSAC_{Low}$ can predict better results when the dense distribution of points is occasionally present in the high outlier scene. These results demonstrate that RLSAC can effectively learn the distribution in point groups to classify inliers and outliers.

\begin{table}[t]\scriptsize 
\centering
\caption{\textbf{The ablation study results} of RLSAC on fundamental matrix estimation.}
\setlength{\tabcolsep}{0.9mm}
\renewcommand\arraystretch{1.1}
\begin{tabular}{c|l||cc|cc}
\toprule
\multirow{2}{*}{}    & \multirow{2}{*}{Method}                           & \multicolumn{2}{c|}{mAA@$10^{\circ}$ ↑} & \multicolumn{2}{c}{Median ($^{\circ}$) ↓}  \\ \cline{3-6} 
                     &                                                   & \textbf{R}        & \textbf{t}        & ${\epsilon_\textbf{R}}$        & ${\epsilon_\textbf{t}}$    \\\hline \hline \noalign{\smallskip}
\multirow{2}{*}{(a)} & Ours (w/o descriptors)                            & 0.702        & 0.568        & 1.400        & 2.963   \\
                     & Ours (full, with 128 DIM descriptors) & \textbf{0.760}        & \textbf{0.622}        & \textbf{0.926}        & \textbf{1.751}       \\ \hline
\multirow{6}{*}{(b)} & \makecell[l]{Ours (with max sampling in training\\ and max sampling in testing)}                & 0.730        & 0.591        & 1.132        & 2.331        \\
                     & \makecell[l]{Ours (with max sampling in training\\ and probabilistic sampling in testing)}                    & 0.706        & 0.531        & 1.415        & 2.893      \\
                     & \makecell[l]{Ours (with probabilistic sampling in training\\ and probabilistic sampling in testing)}                    & 0.720        & 0.581        & 1.150        & 2.508      \\
                     & \makecell[l]{Ours (full, with probabilistic sampling in training\\ and max sampling in testing)}                    & \textbf{0.760}        & \textbf{0.622}        & \textbf{0.926}        & \textbf{1.751}      \\ \hline
\multirow{4}{*}{(c)} & Ours (with N=100 points)                          & 0.727        & 0.588        & 1.216        & 2.464    \\
                     & Ours (with N=200 points)                          & 0.747        & 0.604        & 1.050        & 2.108    \\
                     & Ours (with N=300 points)                          & 0.733        & 0.594        & 1.180        & 2.364    \\
                     & Ours (full, with N=150 points)                    & \textbf{0.760}        & \textbf{0.622}        & \textbf{0.926}        & \textbf{1.751}    \\ \bottomrule
\end{tabular}
\label{tab:ablation}
\vspace{-0.5cm}	
\end{table}

\textbf{Effect of Descriptors:} 
To investigate the effect of descriptors with image semantic features, we conduct experiments as shown in Table \ref{tab:ablation} (a). This suggests that descriptors with semantic features are helpful for RLSAC to effectively learn and sample minimum sets.


\textbf{Different Sampling Approaches:} 
In Table \ref{tab:ablation} (b), four different sampling strategies are compared. The results illustrate that the best performance can be achieved by probabilistic sampling during training and max sampling during testing. This is consistent with the sampling strategy for actions in reinforcement learning. Specifically, probabilistic sampling can provide randomness and exploratory during training, while max sampling can output the optimal strategy during testing to improve performance and efficiency.

\textbf{Number of State Points:} 
In RLSAC, the state points are selected from the top $N$ correspondences sorted by SNN. Different values of $N$ are evaluated in Table \ref{tab:ablation} (c). The best performance is obtained when $N=150$. We count the number of correspondences in images at $SNN<0.8$ recommended in \cite{barath2020ransac}, which is also around 150 points. This is reasonable that a smaller number of points may exclude points, which would solve the better hypothesis. While more points will introduce more noise, making it more difficult for RLSAC to learn effective sampling strategies.

\section{Discussion}
The methods of estimating results directly with neural networks in one shot \cite{poursaeed_deep_2018} are not robust and generalized in the scenes that have not been learned. Moreover, their poor interpretability leads to poor practicality in engineering applications. In contrast, traditional methods based on mathematical theory, such as multiple sampling consensus for model estimation and noise covariance matrix estimation in simultaneous localization and mapping (SLAM), offer clear interpretability and well-defined scopes of application. However, many traditional methods cannot be integrated into a learning-based framework due to their non-differentiability.

Therefore, combining these traditional methods with learning-based methods can provide both interpretability and high performance. RLSAC combines sampling consensus with reinforcement learning to avoid differentiation of sampling while better extracting data features and memory features. In addition, RLSAC can be easily transferred to other sampling consensus tasks due to:
\vspace{-0.1cm}
\begin{itemize}
        \vspace{-0.1cm}
	\item The input data of RLSAC is not limited to coordinates and descriptors. Additional features such as depth estimation information and semantic segmentation information can also be used as input for the policy network.
 
        \vspace{-0.2cm}
	\item RLSAC retains random sampling at the beginning of each episode and each sample result is de-duplicated. This provides basic performance for sampling consensus-based tasks.

         \vspace{-0.2cm}
	\item The rewards in RLSAC are linked with the evaluation of the hypothesis. Therefore, RLSAC can be extended to other robust estimation tasks, which require different evaluation methods. Moreover, RLSAC does not require differentiation of the sampling process, the hypothesis solving process, and the evaluation process.
\end{itemize}
\vspace{-0.1cm}
In this way, RLSAC enables end-to-end learning that incorporates traditional methods. This allows for consistent optimization objectives across modules.

\section{Conclusion}
In this paper, we propose RLSAC, a reinforcement learning enhanced sample consensus framework for end-to-end robust estimation. RLSAC models the sampling consensus process as a reinforcement learning task to achieve end-to-end robust estimation. The basic performance of RLSAC is provided by the initial random sampling in each episode. In RLSAC, a new state transition strategy is designed to effectively extract current and historical information, which can guide RLSAC to explore state space. Furthermore, the inlier ratio of the hypothesis is used as a reward to realize unsupervised policy network learning. In experiments, the 2D line fitting task illustrates that RLSAC is robust to various disturbances and exhibits strong generalization. The visualization of the sampling process shows that RLSAC can progressively explore to better models and continues to explore locally. Additionally, the fundamental matrix estimation task demonstrates that RLSAC outperforms other methods in the complex camera pose estimation problem. Notably, RLSAC is not limited to specific tasks and can easily perform end-to-end learning on various sampling consensus-based tasks. In future work, it is worthwhile to explore the integration of traditional and learning-based methods within this framework, while trying to maintain the strengths of each method.

\section{Acknowledgement}
This work was supported in part by the Natural Science Foundation of China under Grant 62225309, 62073222, U21A20480 and U1913204. The authors gratefully appreciate the contribution of Yiqing Xu from CUMT.

{\small
\bibliographystyle{ieee_fullname}
\bibliography{egbib}
}

\end{document}